\documentclass[11pt]{article}

\usepackage[preprint]{acl}

\usepackage{times}
\usepackage{latexsym}
\usepackage[T1]{fontenc}
\usepackage[utf8]{inputenc}
\usepackage{microtype}
\usepackage{inconsolata}
\usepackage{graphicx}
\usepackage{amsmath}
\usepackage{amssymb}
\usepackage{booktabs}
\usepackage{array}
\usepackage{tabularx}
\usepackage{multirow}
\usepackage{float}
\usepackage{placeins}
\usepackage{needspace}
\usepackage{pgfplots}
\pgfplotsset{compat=1.18}
\newcolumntype{Y}{>{\raggedright\arraybackslash}X}

\title{Smoothing the Top-$k$ Exposure Boundary for Sparse Mixture-of-Experts}

\author{
  \textbf{Yunkai Chai\textsuperscript{1,2}, Tong Zhu\textsuperscript{2},
          Xiaoye Qu\textsuperscript{2}, Xuyang Hu\textsuperscript{2}} \\
  \textbf{Guanjie Chen\textsuperscript{2}, Qipeng Guo\textsuperscript{2},
          Yu Cheng\textsuperscript{3}} \\
  \textnormal{\textsuperscript{1}Shanghai Jiao Tong University, Shanghai, China} \\
  \textnormal{\textsuperscript{2}Shanghai AI Laboratory, Shanghai, China} \\
  \textnormal{\textsuperscript{3}Nanyang Technological University, Singapore} \\
  \textnormal{\texttt{chaiyunkai@sjtu.edu.cn, \{zhutong,quxiaoye,huxuyang,guoqipeng\}@pjlab.org.cn}} \\
  \textnormal{\texttt{chenguanjie@sjtu.edu.cn, yu.cheng@ntu.edu.sg}}
}

\begin{document}
\maketitle

\begin{abstract}

Sparse Mixture-of-Experts models scale parameter capacity efficiently while maintaining a fixed compute budget per token.
However, traditional training paradigms enforce a static choice of top-$k$ experts, which converts a continuous routing distribution into a rigid step function.
This constraint introduces a brittle boundary where highly competitive experts are arbitrarily separated into full-supervision and zero-feedback zones based on minor score fluctuations.
To address this issue, we propose Elastic Expert Routing, which stochastically samples the active expert budget from a localized discrete distribution centered at $k$.
Over multiple training iterations, this mechanism softens the sharp threshold into a gradual probability distribution.
Because the sampling neighborhood remains symmetric, this approach matches the expected computational cost of deterministic training, while preserving the inference budget.
During supervised fine-tuning, elastic routing improves downstream macro-averages on OLMoE-1B-7B and Qwen3-30B-A3B by $+0.84$ and $+2.02$ points, respectively.
In addition, in from-scratch pretraining, it reduces validation LM loss by up to $0.04$ and outperforms the static top-$k$ baseline by $1.6$ points on average across downstream tasks.
Comparisons on Qwen3 fine-tuning and from-scratch pretraining further show higher average scores than Noisy top-$k$, expert FFN dropout, and DefaultMoE.
\end{abstract}

\begin{figure}[t]
\centering
\includegraphics[width=\columnwidth]{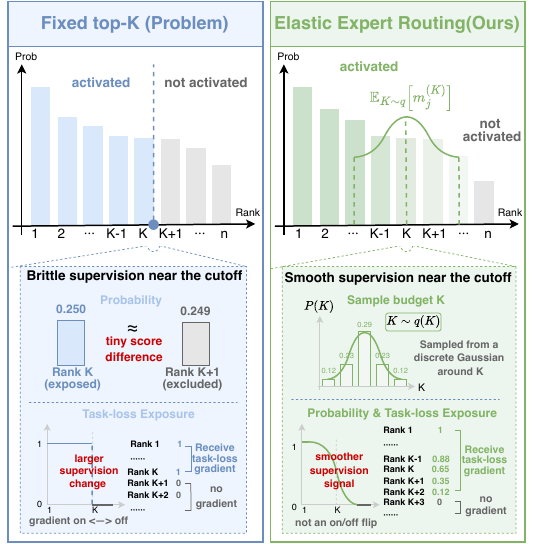}
\caption{
Routing boundaries of static top-$k$ and Elastic Expert Routing.
}
\label{fig:overview_elastic_routing}
\end{figure}

\section{Introduction}
\label{sec:introduction}

Sparse Mixture-of-Experts (MoE) could scale model capacity efficiently by enlarging the number of total experts while conditionally activating a restricted subset of parameters per token \citep{jiang2024mixtralexperts,dai2024deepseekmoe,du2022glamefficientscalinglanguage}.
In the common top-$k$ architecture, a routing network ranks the experts for each incoming token and dispatches that token to the first $k$ candidates~\cite{shazeer2017outrageouslylargeneuralnetworks,fedus2022switchtransformersscalingtrillion}.
Since the choice of $k$ directly serves as a static computational hyperparameter that dictates per-token costs, this design provides an intuitive mechanism for controlling computational requirements, where the compute would stay almost the same while the total model capacity scales~\cite{tian2025greaterleveragescalinglaws,krajewski2024scaling}.
Consequently, a common practice would enforce a uniform value of $k$ throughout both the training and deployment phases.

However, this fixed threshold converts a continuous routing distribution into a rigid step function at the selection threshold~\cite{wang2025remoe}.
This constraint introduces a harsh boundary that restricts model optimization.
As shown in Figure~\ref{fig:overview_elastic_routing}, highly competitive experts with virtually identical routing scores are arbitrarily separated into full-supervision and zero-feedback zones based on minor score fluctuations.
Consequently, an expert ranked precisely inside the cutoff receives full gradient supervision, whereas an adjacent competitor ranked just outside receives zero task-loss feedback.
Because traditional top-$k$ training repeatedly presents the routing network with this rigid boundary, optimization near the cutoff becomes highly brittle.

One intuitive approach to mitigate boundary rigidity is dynamic routing.
Existing adaptive methods vary the number of activated experts dynamically by allocating more capacity to difficult inputs~\cite{guo2025dynamicmixtureexpertsautotuning} or sampling token-specific budgets from routing signals \citep{yue2024adak}.
While this strategy removes the top-$k$ threshold, it may change the deployment budget~\cite{wang2025remoe,huang2024hardertasksneedexperts}, alter the model structure~\cite{zeng2024adamoe,jin2025moe++,team2025longcat}, or lead to suboptimal task performance for extreme efficiency~\cite{lv2026post}, which is hard to be applied on model serving infrastructures.

To address this issue, we propose Elastic Expert Routing. Our framework preserves the standard fixed-budget inference constraints but softens the sharp threshold during training.
Specifically, our method stochastically samples the active expert budget from a localized discrete distribution centered at $k$.
As illustrated in Figure~\ref{fig:overview_elastic_routing} (right), over multiple training iterations, this mechanism statistically transforms the sharp selection threshold into a smooth selection probability distribution.
Consequently, competitive near-boundary experts can be optimized during training.
Because the sampling neighborhood remains symmetric around $k$, this approach matches the expected computational cost of vanilla static top-$k$ training.
Furthermore, the model preserves the standard static top-$k$ routing rule at inference time with zero structural modification or latency overhead.

We evaluate the efficacy of elastic routing across fine-tuning and pretraining settings.
In supervised fine-tuning experiments, we test our method using OLMoE-1B-7B~\cite{muennighoff2025olmoeopenmixtureofexpertslanguage} and Qwen3-30B-A3B backbones \citep{yang2025qwen3technicalreport}.
Our approach consistently improves downstream task performance over static top-$k$ baselines, yielding improvements of $+0.84$ and $+2.02$ points, respectively.
We further validate the foundational utility of budget elasticity during from-scratch pretraining under strict compute-matched and token-matched conditions.
In this setting, elastic routing reduces validation LM loss by up to $0.04$ and outperforms the static top-$k$ baseline by $1.6$ points on average across downstream tasks.
On Qwen3 fine-tuning and from-scratch pretraining, elastic routing also outperforms Noisy top-$k$, expert FFN dropout, and DefaultMoE in average downstream performance under the same static top-8 inference rule.
Subsequent internal diagnostics confirm that boundary smoothing refines expert prioritization rankings and alters routing distribution behaviors within the foundational middle layers of the network.

Our contributions are summarized below:
\begin{itemize}
    \item We formulate static top-$k$ MoE training as a hard rank-exposure boundary
    and introduce Elastic Expert Routing to soften it into a gradual selection probability distribution without modifying inference mechanics.
    \item We demonstrate consistent performance gains during supervised fine-tuning on OLMoE-1B-7B and Qwen3-30B-A3B backbones, as well as during from-scratch pretraining under fair settings.
    \item We provide routing diagnostics and router-swap interventions showing
    that the gains are associated with middle-layer routing changes and improved
    expert prioritization.
\end{itemize}

\section{Preliminary and Related Work}
\label{sec:related_work}

\paragraph{Sparse MoE routing.}
Sparse MoE models replace dense feed-forward blocks with sparsely activated expert modules
\citep{shazeer2017outrageouslylargeneuralnetworks,lepikhin2020gshardscalinggiantmodels,fedus2022switchtransformersscalingtrillion,du2022glamefficientscalinglanguage,zoph2022stmoedesigningstabletransferable,jiang2024mixtralexperts}.
For a token state $h_i$, the router computes routing logits
\begin{equation}
z_i=W_{\mathrm{gate}}h_i,\qquad p_i=\mathrm{softmax}(z_i),
\end{equation}
and under static top-$k$ routing selects
\begin{equation}
S_i^{(k)}=\mathrm{TopK}(p_i,k),\qquad
y_i=\sum_{e\in S_i^{(k)}}\alpha_{i,e}E_e(h_i).
\end{equation}
Fixing $k$ during training and inference creates a hard exposure boundary: only selected experts participate in each token's task-loss computation. The budget therefore shapes both per-token computation and the training exposure distribution.

\paragraph{Routing regularization.}
MoE training commonly relies on auxiliary objectives, capacity constraints, and routing regularization to prevent expert collapse and improve expert utilization \citep{shazeer2017outrageouslylargeneuralnetworks,fedus2022switchtransformersscalingtrillion,zoph2022stmoedesigningstabletransferable}. 
Other work revisits token--expert assignment by changing the assignment rule itself, for example allowing experts to select tokens rather than assigning a fixed number of experts per token \citep{zhou2022mixtureofexpertsexpertchoicerouting}, or introducing alternative selection mechanisms based on BASE layers~\cite{lewis2021baselayerssimplifyingtraining} and differentiable sorting and ranking~\cite{wang2025remoe,hazimeh2021dselectkdifferentiableselectionmixture} strategies.
In contrast, our approach preserves the routing architecture and assignment mechanism, and instead changes the training distribution over the active expert budget.
Noisy top-$k$ perturbs routing scores while retaining a fixed expert count \citep{shazeer2017outrageouslylargeneuralnetworks}; our FFN dropout baseline instead perturbs expert activations. DefaultMoE approximates uncomputed expert outputs with exponential moving averages to provide dense router feedback \citep{panda2025defaultmoe}, whereas elastic routing changes the active expert count and near-boundary exposure.

\paragraph{Dynamic expert routing.}
Recent work relaxes static top-$k$ routing by allowing the number of activated experts to vary across tokens. Existing approaches mainly differ in how dynamic budgets are determined, including threshold-based routing such as cumulative, trainable, and non-linear thresholding \citep{huang2024hardertasksneedexperts,guo2025dynamicmixtureexpertsautotuning,wang2025remoe}, auxiliary proposer modules for predicting token-specific budgets \citep{yue2024adak}, and architectural designs based on low- or zero-computation experts \citep{zeng2024adamoe,jin2025moe++,team2025longcat}.
Matryoshka MoE varies the number of active experts during training to support elastic inference across different budgets \citep{wang2025matryoshka}. Our method instead samples a local, centered budget distribution to improve fixed top-$k$ inference, smoothing rank exposure while preserving the expected active-expert count.

\section{Elastic Expert Routing}
\label{sec:setting}

\subsection{Budget-Neighborhood Training}
\label{sec:budget_neighborhood}

We first describe elastic routing as a change to the training objective. Let $\mathcal{D}$ denote the training distribution and $g(\hat{y},y)$ denote the task loss. For an input $x$, let
$f_{\theta}(x;\boldsymbol{\kappa})$ denote the
MoE forward pass under a budget schedule $\boldsymbol{\kappa}$, where
$\kappa_{i,\ell}$ denotes the number of experts used by token $i$ in MoE layer
$\ell$. The fixed-budget baseline uses the constant schedule
$\boldsymbol{k}_0$, with every entry equal to the target budget $k_0$, and
optimizes
\begin{equation}
\begin{aligned}
\mathcal{L}_{\mathrm{fixed}}(\theta)
&=
\mathbb{E}_{(x,y)\sim\mathcal{D}}
\left[
g(f_{\theta}(x;\boldsymbol{k}_0),y)
\right],\\
\boldsymbol{k}_0&\equiv(k_0,\ldots,k_0).
\end{aligned}
\end{equation}
This objective trains at a single budget configuration. Elastic routing instead samples
a schedule $\boldsymbol{\kappa}$ from a distribution $Q$ centered at the
target budget and optimizes
\begin{equation}
\mathcal{L}_{\mathrm{elastic}}(\theta)
=
\mathbb{E}_{(x,y)\sim\mathcal{D}}
\mathbb{E}_{\boldsymbol{\kappa}\sim Q}
\left[
g(f_{\theta}(x;\boldsymbol{\kappa}),y)
\right].
\end{equation}
The fixed-budget objective is recovered when $Q$ degenerates to a point mass at
$\boldsymbol{k}_0$. In the remainder of this section, $Q$ is local around
$k_0$ and centered in expectation,
$\mathbb{E}_{\boldsymbol{\kappa}\sim Q}[\kappa_{i,\ell}]=k_0$ for each token
and layer, so elastic routing changes the training-time budget distribution
while preserving the expected active-expert count.
Section~\ref{sec:boundary_smoothing} formalizes the resulting
rank-neighborhood smoothing.

\subsection{Instantiating the Budget Neighborhood}
\label{sec:budget_distribution}
We now specify the budget distribution used for elastic routing.
An $a$--$b$ schedule restricts the sampled budget to
$\mathcal{K}_{a:b}=\{a,\ldots,b\}$ and assigns probability
\begin{equation}
\begin{aligned}
q_{a:b}(r)
&=
\frac{1}{Z_{a:b}}
\exp \left(- \frac{(r-k_0)^2}{2\sigma^2} \right),
\quad r\in\mathcal{K}_{a:b},\\
Z_{a:b}
&=
\sum_{s=a}^{b}
\exp \left(- \frac{(s-k_0)^2}{2\sigma^2} \right).
\end{aligned}
\end{equation}
Here $r$ denotes a candidate active-expert budget, $k_0$ is the center, and
$\sigma$ controls the neighborhood width.
Unless otherwise stated, all elastic schedules in our experiments use
$\sigma=1.5$. For each token $i$ and MoE layer $\ell$, we sample
$\kappa_{i,\ell}\sim q_{a:b}$ and dispatch the token to the
top-$\kappa_{i,\ell}$ experts under the current router ranking.
Experts outside the sampled prefix are masked out, and
load-balancing statistics are computed over the realized dispatched assignments.

\subsection{Rank-Neighborhood Smoothing}
\label{sec:boundary_smoothing}

We formalize boundary smoothing through rank-exposure accounting. Conditioned on
the current router order for representation $h$, let $\pi_j(h)$ denote the expert
at rank $j$ and define the top-$r$ prefix mask
\begin{equation}
m_j^{(r)}=\mathbf{1}[j\leq r].
\end{equation}
For a scalar budget distribution $q$, the task-loss exposure probability of
rank $j$ is
\begin{equation}
\begin{aligned}
a_q(j)
&=
\mathbb{E}_{K\sim q}\!\left[m_j^{(K)}\right]\\
&=
\Pr(K\geq j)
=
\sum_{r=j}^{K_{\max}}q(r).
\end{aligned}
\end{equation}
Static top-$k_0$ training corresponds to the point-mass case
$a_{\delta_{k_0}}(j)=\mathbf{1}[j\leq k_0]$, whereas elastic routing replaces
this step mask with the survival function of $q$.

The budget distribution is the discrete derivative of the exposure curve:
\begin{equation}
a_q(j)-a_q(j+1)=q(j),\qquad a_q(M+1)=0.
\end{equation}
Thus the smoothing effect is local in router rank space, not a differentiable relaxation
of top-$K$ selection. If $\mathrm{supp}(q)=[K_{\min},K_{\max}]$, then ranks
$j\leq K_{\min}$ are always exposed, ranks
$K_{\min}<j\leq K_{\max}$ have fractional exposure, and ranks
$j>K_{\max}$ are never exposed.

The same accounting yields expected exposure conservation:
\begin{equation}
\sum_{j=1}^{M}a_q(j)
=
\sum_{j=1}^{M}\Pr(K\geq j)
=
\mathbb{E}[K].
\end{equation}
When $q$ is exactly centered at the deployment budget,
$\mathbb{E}[K]=k_0$, elastic routing preserves the expected active-expert
count while changing which boundary ranks enter the task-loss path.
Moreover,
\begin{equation}
\sum_{j>k_0} a_q(j)
=
\sum_{j\leq k_0} \bigl(1-a_q(j)\bigr)
=
\frac{1}{2}\mathbb{E}|K-k_0|,
\end{equation}
so the exposure above the cutoff is balanced by exposure from ranks at or below it.

The schedule width controls how much conserved exposure shifts across the deployment cutoff: narrow neighborhoods transfer little exposure, while wider
neighborhoods shrink the always-on core and include lower-ranked candidates.

\section{Experiments}
\label{sec:exp_setup}

\subsection{Setup}
\label{sec:experimental_protocol}
\paragraph{Models and data.}
We evaluate elastic routing in two supervised fine-tuning (SFT) settings and one from-scratch pretraining setting. 
For SFT, we fine-tune OLMoE-1B-7B-0924 on Dolci-Instruct-SFT \citep{olmo2026olmo3} and Qwen3-30B-A3B-Base on Infinity-Instruct 7M Core \citep{li2025infinityinstructscalinginstruction}. 
For pretraining experiments, we train 1.42B-A365M MoE models with 6e19 FLOPs budget.
The model structure, optimal learning rate, and global batch size are guided by the scaling laws of \citet{tian2025greaterleveragescalinglaws} and \citet{hoffmann2022training}.
The model has 128 experts per layer and is trained on FineWeb-Edu \citep{penedo2024finewebdatasetsdecantingweb} for 26.25B tokens.

\paragraph{Routing conditions.}
Each setting compares static \texttt{top-8} training with token-level elastic schedules \texttt{7--9}, \texttt{6--10}, and \texttt{5--11}. Elastic budgets are sampled from the discrete Gaussian neighborhood in Section~\ref{sec:budget_distribution}, with $\sigma=1.5$ and center $k_0=8$. 
Since the schedules are centered at \texttt{top-8}, they match the baseline in expected active-expert count and expected expert-side training FLOPs. All main evaluations use static \texttt{top-8} inference; training cost and system profiling results are reported in Appendix~\ref{app:measured_cost} (Tables~\ref{tab:app_measured_cost} and~\ref{tab:app_system_profile}).

\paragraph{Evaluation.}
For SFT, we report Avg(9) over a nine-task suite covering
\textbf{Knowledge}: MMLU \citep{hendrycks2021measuringmassivemultitasklanguage}, AGIEval (En)
\citep{zhong2023agievalhumancentricbenchmarkevaluating}, GPQA (Main)
\citep{rein2023gpqagraduatelevelgoogleproofqa},
\textbf{Reasoning}: BBH \citep{suzgun2022challengingbigbenchtaskschainofthought}, \textbf{Mathematics}: GSM8K \citep{cobbe2021trainingverifierssolvemath}, MATH
\citep{hendrycksmath2021},
\textbf{Coding}: MBPP \citep{austin2021programsynthesislargelanguage},
and \textbf{Instruction Following \& Alignment}: IFEval
\citep{zhou2023instructionfollowingevaluationlargelanguage}, TruthfulQA
\citep{lin2022truthfulqameasuringmodelsmimic}.
For pretraining, we report averaged results on {LAMBADA} \citep{paperno2016lambadadatasetwordprediction};
{HellaSwag~\cite{zellers2019hellaswagmachinereallyfinish}}, {PIQA~\cite{bisk2019piqareasoningphysicalcommonsense}}, {WinoGrande~\cite{sakaguchi2019winograndeadversarialwinogradschema}},{CommonsenseQA~\cite{talmor2019commonsenseqaquestionansweringchallenge}},
{SciQ~\cite{welbl2017crowdsourcingmultiplechoicescience}}, {OpenBookQA~\cite{mihaylov2018suitarmorconductelectricity}}, {ARC~\cite{clark2018thinksolvedquestionanswering}},
{MMLU} \citep{hendrycks2021measuringmassivemultitasklanguage},
and {LogiQA} \citep{liu2020logiqachallengedatasetmachine}.
We use standard task metrics through lm-evaluation-harness \citep{lm-eval-harness}.
See Appendix~\ref{app:reproducibility} for details.

\subsection{Supervised Fine-Tuning Results}
\label{sec:main_results}
Table~\ref{tab:main_results} reports the SFT results under static
\texttt{top-8} inference. Comparisons are made within each backbone: every
elastic checkpoint shares the same data, optimizer, evaluation protocol, and
decoding setup as its static \texttt{top-8} counterpart, and differs only in the
training-time routing schedule.

\begin{table*}[t]
\centering
\small
\setlength{\tabcolsep}{4.5pt}
\resizebox{\textwidth}{!}{
\begin{tabular}{llrrrrrrrr}
\toprule
\multirow{2}{*}{\textbf{Task}} & \multirow{2}{*}{\textbf{Metric}}
& \multicolumn{4}{c}{\textbf{OLMoE-1B-7B}}
& \multicolumn{4}{c}{\textbf{Qwen3-30B-A3B}} \\
\cmidrule(lr){3-6}\cmidrule(lr){7-10}
& & \texttt{\textbf{6--10}} & \texttt{\textbf{5--11}} & \texttt{\textbf{7--9}} & \texttt{\textbf{top-8}}
& \texttt{\textbf{6--10}} & \texttt{\textbf{5--11}} & \texttt{\textbf{7--9}} & \texttt{\textbf{top-8}} \\
\midrule
MMLU & Accuracy & \textbf{51.25} & 50.91 & 50.14 & 50.47 & \textbf{78.49} & 78.40 & 78.17 & 78.24 \\
AGIEval (En) & Accuracy & \textbf{23.96} & 23.44 & 22.48 & 23.29 & 52.28 & 51.90 & \textbf{52.44} & 51.74 \\
GPQA (Main) & Accuracy & 26.34 & \textbf{26.79} & 26.12 & 25.00 & \textbf{39.06} & 38.39 & 37.28 & 37.72 \\
BBH & Exact Match & 35.28 & \textbf{36.05} & 34.54 & 34.86 & \textbf{63.29} & 57.69 & 55.68 & 49.29 \\
GSM8K & Exact Match & \textbf{21.46} & 21.38 & 19.86 & 20.09 & 82.03 & \textbf{82.94} & 82.71 & 81.43 \\
MATH & Exact Match & 6.44 & 6.42 & \textbf{6.52} & 6.40 & \textbf{42.36} & 37.78 & 42.18 & 41.24 \\
MBPP & Pass@1 & 25.40 & 25.00 & \textbf{25.60} & 24.20 & 74.00 & \textbf{76.00} & 73.40 & 73.80 \\
IFEval & Accuracy & \textbf{51.02} & 48.98 & 50.46 & 50.46 & 42.51 & \textbf{43.44} & 40.48 & 41.22 \\
TruthfulQA & MC2 Accuracy & 41.58 & 41.15 & \textbf{41.60} & 40.32 & 51.93 & 52.65 & 52.28 & \textbf{53.14} \\
\midrule
\multicolumn{2}{c}{Average} & \textbf{31.41} & 31.12 & 30.81 & 30.57 & \textbf{58.44} & 57.69 & 57.18 & 56.42 \\
\bottomrule
\end{tabular}}
\caption{Main benchmark results across two sparse MoE backbones. Avg(9) is the
unweighted average over the nine reported tasks. Bold
marks the best method within each backbone. All elastic checkpoints are evaluated with the same static
top-8 inference rule as their corresponding \texttt{top-8} baseline.}
\label{tab:main_results}
\end{table*}

\begin{table}[t]
\centering
\resizebox{\linewidth}{!}{
\begin{tabular}{lrrrr}
\toprule
\textbf{Dataset / Metric} & \textbf{\texttt{top-8}} & \textbf{\texttt{6--10}} & \textbf{\texttt{7--9}} & \textbf{\texttt{5--11}} \\
\midrule
Validation LM loss $\downarrow$ & 2.6465 & \textbf{2.6085} & 2.6160 & 2.6369 \\
LAMBADA PPL $\downarrow$ & 54.2 & \textbf{42.7} & 44.9 & 48.7 \\
LAMBADA & 31.3 & \textbf{33.1} & 32.2 & 31.7 \\
HellaSwag & 41.9 & \textbf{43.7} & 43.5 & 42.2 \\
PIQA & 68.4 & \textbf{68.6} & 68.3 & 66.9 \\
WinoGrande & 49.3 & 52.6 & 53.0 & \textbf{53.4} \\
CommonsenseQA & 19.4 & 20.1 & 19.8 & \textbf{20.2} \\
SciQ & 81.4 & 82.0 & \textbf{82.2} & 81.9 \\
OpenBookQA & 31.0 & 33.8 & \textbf{35.2} & 34.0 \\
ARC-Easy & 53.7 & 55.1 & 55.3 & \textbf{55.6} \\
ARC-Challenge & 28.5 & 29.9 & \textbf{31.1} & 29.7 \\
MMLU & 23.1 & \textbf{26.1} & 24.1 & 23.3 \\
LogiQA & 26.3 & \textbf{27.2} & 26.3 & 26.9 \\
\midrule
\textbf{Average (excl. loss/PPL)} & 41.3 & \textbf{42.9} & 42.8 & 42.4 \\
\bottomrule
\end{tabular}
}
\caption{From-scratch pretraining validation under matched architecture, data,
token budget, and expected active expert count. PPL denotes perplexity, where
lower is better. Validation LM loss is the logged validation loss at step 25,034.
All other metrics except PPL are percentages; Avg excludes loss and perplexity.}
\label{tab:pretrain_validation}
\end{table}

\paragraph{Elastic routing improves static \texttt{top-8} SFT.} On OLMoE, all three elastic schedules outperform the static
\texttt{top-8} baseline, with \texttt{6--10} increasing Avg(9) from 30.57 to
31.41. On Qwen3-30B-A3B, \texttt{6--10} raises Avg(9) from 56.42 to 58.44,
and the other elastic schedules also remain above the baseline. The improvement
is broad: an elastic schedule obtains the best score on all nine OLMoE tasks
and on eight of nine Qwen3 tasks.

\paragraph{Moderate interval gives the strongest aggregate result.} The
\texttt{6--10} schedule gives the best Avg(9) for both backbones, whereas the
narrower \texttt{7--9} and wider \texttt{5--11} schedules are beneficial but
not consistently strongest. This pattern agrees with the rank-exposure view
in Section~\ref{sec:boundary_smoothing} that useful smoothing should expose experts
near the cutoff threshold without allocating excessive routing probability to low-ranked, irrelevant candidates.

\subsection{From-Scratch Pretraining Validation}
\label{sec:pretrain_validation}
To test whether the effect could be generalized to pretraining, we conduct experiments to train four MoE baselines from scratch.
The downstream task results are listed in Table~\ref{tab:pretrain_validation}.

\paragraph{Elastic routing improves from-scratch pretraining under matched compute.}
All three elastic schedules improve the average validation score over static
\texttt{top-8}: \texttt{6--10} raises the average from 41.3 to 42.9, while
\texttt{7--9} and \texttt{5--11} reach 42.8 and 42.4, also significantly beyond baseline.

\paragraph{Moderate interval remains strongest in pretraining.} The
\texttt{6--10} schedule gives both the best average score and the lowest LAMBADA perplexity, reducing perplexity from 54.2 to 42.7. Elastic \texttt{6--10} also achieves the lowest validation LM loss, reducing it from 2.6465 to 2.6085 under static top-8 evaluation. This is consistent with the SFT results, supporting elastic routing as a training-time improvement for static \texttt{top-8} MoE models. Its repeated advantage across SFT and pretraining suggests that this neighborhood size captures a robust local smoothing regime rather than a setting-specific fluctuation.

\subsection{Comparison with Other MoE Training Methods}
\label{sec:training_method_comparison}
\textbf{We compare elastic routing with three alternative MoE training methods under matched settings.} The baselines are Noisy top-$k$ \citep{shazeer2017outrageouslylargeneuralnetworks}, expert FFN dropout, and DefaultMoE \citep{panda2025defaultmoe}. Within each Qwen3-30B-A3B SFT or 1.42B MoE pretraining setting, all methods use the same backbone, training data, and optimization budget, and are evaluated with static top-8 routing.

\textbf{Elastic \texttt{6--10} achieves the highest average in both settings.} It reaches 58.44 on the nine-task SFT suite and 42.9 on the eleven-task pretraining suite, while also achieving the lowest pretraining validation LM loss of 2.6085 (Table~\ref{tab:training_method_comparison}). Appendix~\ref{app:training_method_comparison} provides the baseline configurations and full task-level results in Tables~\ref{tab:baseline_tasks_sft} and~\ref{tab:baseline_tasks_pt}.

\begin{table}[t]
\centering
\small
\setlength{\tabcolsep}{4pt}
\resizebox{\linewidth}{!}{
\begin{tabular}{lrrr}
\toprule
\multirow{2}{*}{\textbf{Method}} & \textbf{Qwen3 SFT} & \multicolumn{2}{c}{\textbf{Pretraining}} \\
\cmidrule(lr){2-2}\cmidrule(lr){3-4}
& \textbf{Avg $\uparrow$} & \textbf{Avg $\uparrow$} & \textbf{LM loss $\downarrow$} \\
\midrule
Static top-8 & 56.42 & 41.3 & 2.6465 \\
Noisy top-$k$ & 57.03 & 42.1 & 2.6252 \\
Expert FFN dropout & 57.72 & 40.1 & 2.6671 \\
DefaultMoE & 57.63 & 42.0 & 2.6399 \\
Elastic \texttt{6--10} & \textbf{58.44} & \textbf{42.9} & \textbf{2.6085} \\
\bottomrule
\end{tabular}}
\caption{Comparison with other MoE training methods under static top-8 inference. SFT Avg covers nine tasks; pretraining Avg covers eleven accuracy metrics, excluding loss and perplexity. LM loss is measured on the pretraining validation set.}
\label{tab:training_method_comparison}
\end{table}

\subsection{Depth-Localized Routing Changes}
\label{sec:mechanistic_analysis}
\label{sec:layerwise_motivation}
\label{sec:routing_diagnostics}

We examine where elastic training changes the router across depth and whether restricting elasticity to these layers improves static top-8 inference.

\begin{figure}[!tbp]
\centering
\begin{tikzpicture}
\begin{axis}[
    width=\columnwidth,
    height=4.7cm,
    ybar,
    bar width=5.2pt,
    bar shift=0pt,
    xlabel={Layer},
    ylabel={Mean JS ($\times 10^{-3}$)},
    xmin=-0.55, xmax=15.55,
    ymin=0, ymax=3.05,
    xtick={0,3,7,11,15},
    minor xtick={1,2,4,5,6,8,9,10,12,13,14},
    ytick={0,1,2,3},
    grid=major,
    major grid style={draw=black!9},
    minor grid style={draw=none},
    axis line style={draw=black!55, line width=0.45pt},
    tick style={draw=black!55, line width=0.45pt},
    ticklabel style={font=\scriptsize},
    label style={font=\small},
    axis x line*=bottom,
    axis y line*=left,
    enlarge x limits=false,
    clip=false,
]
\path[fill=teal!7] (axis cs:2.55,0) rectangle (axis cs:7.45,3.05);
\addplot[draw=black!30, fill=black!18, forget plot] coordinates {
(1,1.691) (2,1.572) (8,1.760) (9,1.524)
(10,1.306) (11,1.375) (12,1.336) (13,1.458) (14,1.495)
};
\addplot[draw=teal!55!black, fill=teal!55, forget plot] coordinates {
(3,2.470) (4,2.327) (5,1.980) (6,2.293) (7,2.011)
};
\addplot[draw=black!50, fill=black!36, forget plot] coordinates {
(0,2.209) (15,2.764)
};
\draw[teal!55!black, line width=0.55pt] (axis cs:3,2.78) -- (axis cs:7,2.78);
\draw[teal!55!black, line width=0.55pt] (axis cs:3,2.68) -- (axis cs:3,2.78);
\draw[teal!55!black, line width=0.55pt] (axis cs:7,2.68) -- (axis cs:7,2.78);
\node[font=\scriptsize, text=teal!45!black, anchor=south] at (axis cs:5,2.80) {L3--L7 block};
\end{axis}
\end{tikzpicture}
\caption{Layerwise diagnostic comparing \texttt{6--10} against static
\texttt{top-8}. We plot mean pairwise JS divergence across tasks. The shaded
region marks the internal block L3--L7 used by the layerwise variants.}
\label{fig:layerwise_diagnostic}
\end{figure}
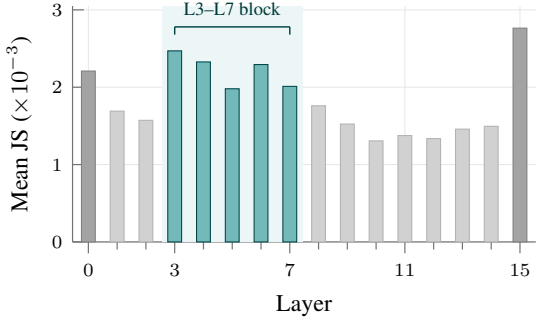

\paragraph{Elastic routing changes the router most strongly in the middle layers.}
We compare \texttt{6--10} and static \texttt{top-8} under the same static top-8 evaluation rule. For each task and layer, we compute the Jensen--Shannon (JS) divergence between the two routers' per-token expert probability distributions. Figure~\ref{fig:layerwise_diagnostic} shows the mean divergence across tasks. The largest non-boundary divergence forms a contiguous L3--L7 block. Although L0 and L15 also exhibit elevated divergence, we exclude these boundary layers from the diagnostic block to separate the internal routing changes from input adaptation and final representation readout.

\paragraph{The depth-restricted diagnostic supports the middle-layer
interpretation.} We apply the \texttt{6--10} rule only to L3--L7, keeping all other layers at static top-8, and use L8--L12 as a later-layer control. Table~\ref{tab:layer_region_ablation} reports the task-level results. Restricting elasticity to L3--L7 reaches Avg(9) 31.50, compared with 31.41 for all-layer \texttt{6--10} and 30.57 for static top-8. The L8--L12 variant also improves over the static baseline, but reaches only 30.94. These results identify L3--L7 as a beneficial diagnostic region, without establishing it as the optimal layer subset.

\begin{table*}[!t]
\centering
\normalsize
\setlength{\tabcolsep}{2pt}
\renewcommand{\arraystretch}{1.18}
\begin{tabular*}{\textwidth}{@{\extracolsep{\fill}}lrrrrrrrrrrr@{}}
\toprule
\textbf{Variant}
& \textbf{MMLU} & \textbf{AGI} & \textbf{GPQA} & \textbf{BBH}
& \textbf{GSM} & \textbf{Min} & \textbf{MBPP} & \textbf{IFE}
& \textbf{TQA} & \textbf{Avg} & $\boldsymbol{\Delta}$ \\
\midrule
\texttt{top-8}
& 50.47 & 23.29 & 25.00 & 34.86 & 20.09 & 6.40 & 24.20 & 50.46 & 40.32 & 30.57 & -- \\
\texttt{6--10}
& \textbf{51.25} & 23.96 & 26.34 & \textbf{35.28} & 21.46 & \textbf{6.44}
& 25.40 & \textbf{51.02} & 41.58 & 31.41 & +0.84 \\
\addlinespace[0.5pt]
\texttt{L3--L7}
& 50.56 & \textbf{24.17} & 27.68 & 35.11 & \textbf{21.91} & 6.16
& \textbf{26.00} & 50.09 & \textbf{41.81} & \textbf{31.50} & \textbf{+0.93} \\
\texttt{L8--L12}
& 50.90 & 23.39 & \textbf{28.12} & 34.77 & 18.65 & 6.26
& 25.60 & 49.72 & 41.00 & 30.94 & +0.37 \\
\bottomrule
\end{tabular*}
\caption{Task-level depth-region ablation on OLMoE. All methods are evaluated
with static top-8 inference. \texttt{L3--L7} corresponds to
an exploratory diagnostic-motivated variant.
Only the middle diagnostic region
uses the \texttt{6--10} training neighborhood, while all other layers remain
static at top-8.
\texttt{L8--L12} applies the same elastic rule to a later block as a
depth-control experiment. Bold marks the best value in each task or summary
column.}
\label{tab:layer_region_ablation}
\end{table*}

\subsection{Expert-Selection Diagnostics and Router Intervention}
\label{sec:expert_choice_router_intervention}

We examine how elastic training changes discrete expert selection, and
whether the learned routing policy contributes to the observed gain. The first
diagnostic experiment compares selected experts under static top-8 evaluation, and the second one uses an evaluation-time router swap to intervene on the routing policy while holding the target model otherwise fixed.

\begin{figure}[t]
\centering
\begin{tikzpicture}
\begin{axis}[
    width=\columnwidth,
    height=5.0cm,
    xlabel={Layer},
    ylabel={Disagreement rate (\%)},
    xmin=-0.25, xmax=15.35,
    ymin=0, ymax=20.5,
    xtick={0,3,7,11,15},
    minor xtick={1,2,4,5,6,8,9,10,12,13,14},
    ytick={0,5,10,15,20},
    grid=major,
    minor grid style={draw=none},
    major grid style={draw=black!9},
    axis line style={draw=black!55, line width=0.45pt},
    tick style={draw=black!55, line width=0.45pt},
    ticklabel style={font=\scriptsize},
    label style={font=\small},
    legend columns=1,
    legend style={
        at={(0.98,0.98)},
        anchor=north east,
        fill=white,
        fill opacity=0.92,
        text opacity=1,
        draw=black!12,
        rounded corners=1pt,
        font=\scriptsize,
        inner sep=2pt,
        row sep=1pt
    },
    clip=false,
]
\addplot+[
    color=blue!70!black,
    line width=1.05pt,
    mark=*,
    mark size=1.75pt,
    mark options={fill=white, draw=blue!70!black, line width=0.65pt}
] coordinates {
(0,17.50) (1,17.90) (2,14.87) (3,17.96)
(4,16.97) (5,15.58) (6,14.37) (7,14.19)
(8,14.87) (9,11.22) (10,10.42) (11,11.87)
(12,10.00) (13,9.65) (14,10.03) (15,10.54)
};
\addlegendentry{$D_1$}
\addplot+[
    color=orange!85!black,
    line width=1.0pt,
    mark=square*,
    mark size=1.65pt,
    mark options={fill=white, draw=orange!85!black, line width=0.6pt}
] coordinates {
(0,10.83) (1,11.03) (2,10.27) (3,10.22)
(4,9.76) (5,8.56) (6,7.48) (7,6.49)
(8,5.72) (9,4.69) (10,4.49) (11,4.44)
(12,4.03) (13,5.09) (14,4.81) (15,5.34)
};
\addlegendentry{$D_8$}
\addplot[blue!70!black, dashed, line width=0.6pt, forget plot] coordinates {(-0.25,13.62) (15.1,13.62)};
\node[font=\scriptsize, text=blue!70!black, fill=white, inner sep=1pt, anchor=east] at (axis cs:15.15,13.62) {13.62 avg};
\addplot[orange!85!black, dashed, line width=0.6pt, forget plot] coordinates {(-0.25,7.08) (15.1,7.08)};
\node[font=\scriptsize, text=orange!85!black, fill=white, inner sep=1pt, anchor=east] at (axis cs:15.15,7.08) {7.08 avg};
\end{axis}
\end{tikzpicture}
\caption{Per-layer expert-selection disagreement between \texttt{6--10} and
static \texttt{top-8}. Top-1 disagreement exceeds normalized top-8 set
disagreement, indicating changes in expert priority and primary assignment
within a largely preserved candidate set.}
\label{fig:expert_identity_disagreement}
\end{figure}
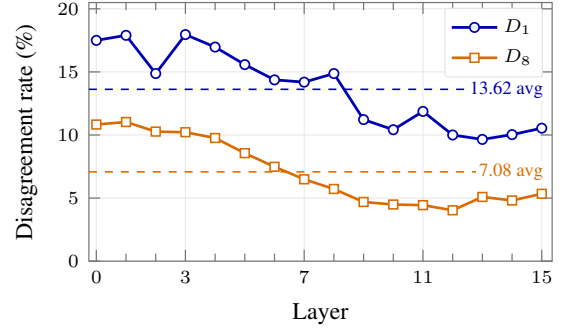

\paragraph{Expert Selection Disagreement.}
We sample 100 evaluation instances from each of our 9 downstream benchmarks to form a representative validation token set $t \in \{1, 2, \dots, T\}$. 
For a given layer $\ell$ and token $t$, let $e^{\text{elastic}}_{\ell,t}$ and $e^{\text{static}}_{\ell,t}$ denote the top-1 expert chosen by each configuration, respectively. 
Similarly, let $S^{\text{elastic}}_{\ell,t}$ and $S^{\text{static}}_{\ell,t}$ represent their respective active top-8 expert sets. 
We define two layer-averaged disagreement ratios, $D_1$ and $D_8$:

\begin{equation}
D_1 = \frac{1}{L \cdot T} \sum_{\ell=1}^{L} \sum_{t=1}^{T} \mathbf{1}\left[ e^{\text{elastic}}_{\ell,t} \neq e^{\text{static}}_{\ell,t} \right],
\end{equation}
\begin{equation}
D_8 = \frac{1}{L \cdot T} \sum_{\ell=1}^{L} \sum_{t=1}^{T} \left( 1 - \frac{\left| S^{\text{elastic}}_{\ell,t} \cap S^{\text{static}}_{\ell,t} \right|}{8} \right),
\end{equation}
where $L$ denotes the total number of layers. 
$D_1 \in [0, 1]$ measures the fraction of top-1 assignments that change, while $D_8 \in [0, 1]$ measures the average fraction of differing experts in the top-8 sets.

Figure~\ref{fig:expert_identity_disagreement} shows that \textbf{elastic routing primarily refines expert rankings rather than altering the global composition of the active expert pool.}
Aggregated across the nine benchmark tasks,
Figure~\ref{fig:expert_identity_disagreement} shows that \texttt{6--10}
changes the top-1 routed expert for 13.62\% of tokens on average, while only
7.08\% of the selected top-$8$ set differs, about 0.57 experts out of 8.

\paragraph{Router-Swap Intervention.}
We swap routers at inference to test their contribution to the gain, keeping all other target parameters fixed.

Table~\ref{tab:router_swap} shows that \textbf{optimal performance gains require the mutual alignment through joint router--expert co-adaptation.}
Replacing the static \texttt{top-8} router with the
\texttt{6--10} router improves the static checkpoint, while replacing the
elastic router with the static router reduces the elastic checkpoint. Neither swap recovers the full \texttt{6--10} performance.

\begin{table}[t]
    \centering
    \begin{tabular}{lllc}
        \toprule
        \textbf{Condition} & \textbf{Source} & \textbf{Target} & \textbf{Avg.} \\
        \midrule
        \multicolumn{4}{l}{\textit{Target model: \texttt{6--10}}} \\
        Original & \texttt{6--10} & \texttt{6--10} & 31.41 \\
        Swapped  & \texttt{top-8} & \texttt{6--10} & 31.29 \\
        \midrule
        \multicolumn{4}{l}{\textit{Target model: \texttt{top-8}}} \\
        Original & \texttt{top-8} & \texttt{top-8} & 30.57 \\
        Swapped  & \texttt{6--10} & \texttt{top-8} & 30.86 \\
        \bottomrule
    \end{tabular}
    \caption{Router-swap intervention with the remaining target model parameters fixed and only the router replaced.}
    \label{tab:router_swap}
\end{table}

\subsection{Performance Under Different Top-$k$ Values}
\label{sec:inference_k_boundary}
\label{sec:inference_k_scan}

Since the main results focus on eight active experts, we further examine whether the gains persist when the number of active experts changes. Elastic routing is designed to expose training to a local range of expert-count
configurations, allowing the model to learn multiple nearby patterns of
expert cooperation instead of only one static top-8 composition. If this
diversity is useful, its effect should not be limited to the exact eight-expert setting used in the main evaluation; it should also appear when
inference uses other expert counts.

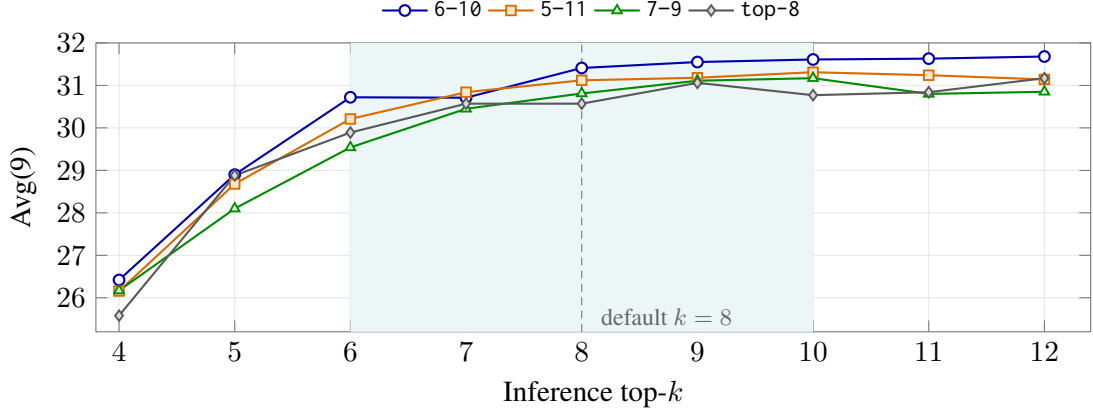
\begin{figure*}[t]
\centering
\begin{tikzpicture}
\begin{axis}[
    width=0.92\textwidth,
    height=5.4cm,
    xlabel={Inference top-$k$},
    ylabel={Avg(9)},
    xmin=3.8, xmax=12.4,
    ymin=25.2, ymax=32.0,
    xtick={4,5,6,7,8,9,10,11,12},
    ytick={26,27,28,29,30,31,32},
    grid=major,
    major grid style={draw=black!10},
    axis line style={draw=black!55},
    tick style={draw=black!55},
    legend columns=4,
    legend style={
        at={(0.5,1.04)},
        anchor=south,
        fill=none,
        draw=none,
        font=\small
    },
    clip=false,
]
\addplot[draw=none, fill=teal!7, forget plot] coordinates {(6,25.2) (10,25.2) (10,32.0) (6,32.0)} \closedcycle;
\addplot[black!55, densely dashed, forget plot] coordinates {(8,25.2) (8,32.0)};
\node[font=\small, text=black!60, anchor=west] at (axis cs:8.08,25.55) {default $k=8$};
\addplot+[
    color=blue!70!black,
    thick,
    mark=*,
    mark size=2.0pt,
    mark options={fill=white, draw=blue!70!black, line width=0.8pt}
] coordinates {
(4,26.42) (5,28.90) (6,30.72) (7,30.71) (8,31.41)
(9,31.55) (10,31.61) (11,31.63) (12,31.68)
};
\addlegendentry{\texttt{6--10}}
\addplot+[
    color=orange!85!black,
    thick,
    mark=square*,
    mark size=1.8pt,
    mark options={fill=orange!20, draw=orange!85!black, line width=0.7pt}
] coordinates {
(4,26.16) (5,28.68) (6,30.21) (7,30.84) (8,31.12)
(9,31.18) (10,31.31) (11,31.24) (12,31.14)
};
\addlegendentry{\texttt{5--11}}
\addplot+[
    color=green!55!black,
    thick,
    mark=triangle*,
    mark size=2.1pt,
    mark options={fill=green!18, draw=green!55!black, line width=0.7pt}
] coordinates {
(4,26.17) (5,28.10) (6,29.54) (7,30.45) (8,30.81)
(9,31.11) (10,31.17) (11,30.80) (12,30.85)
};
\addlegendentry{\texttt{7--9}}
\addplot+[
    color=black!65,
    thick,
    mark=diamond*,
    mark size=2.0pt,
    mark options={fill=black!15, draw=black!65, line width=0.7pt}
] coordinates {
(4,25.58) (5,28.88) (6,29.89) (7,30.57) (8,30.57)
(9,31.06) (10,30.77) (11,30.84) (12,31.17)
};
\addlegendentry{\texttt{top-8}}
\end{axis}
\end{tikzpicture}
\caption{
OLMoE SFT results at different inference top-$k$ values. 
}
\label{fig:inference_k_scan}
\end{figure*}

\noindent
\textbf{Elastic routing improves performance beyond the eight-expert inference setting.} Figure~\ref{fig:inference_k_scan} shows that \texttt{6--10} is the
strongest schedule at the default $k=8$ point and remains above the static
\texttt{top-8} baseline throughout the scanned range. The narrower
\texttt{7--9} and wider \texttt{5--11} schedules also improve some parts of the
curve, but neither gives the same consistently strong profile. This supports the view that training over a moderate range of expert counts
helps the model learn more robust expert cooperation patterns under several nearby
settings, rather than only optimizing the single eight-expert setting.

\subsection{Ablations and Budget Controls}
\label{sec:ablations_controls}

The preceding analyses characterize where elastic routing changes routing and how the learned ordering behaves around the deployment cutoff. We further isolate the effect of budget design by varying the probability shape within a fixed local neighborhood and comparing against simpler alternatives that increase active experts without smoothing the \texttt{top-8} boundary.

\paragraph{Default boundary-smoothing shape performs best.}
The main experiments use $\sigma=1.5$ for all elastic schedules. We fix the \texttt{6--10} support and vary only $\sigma$: smaller values concentrate probability around the target cutoff, while larger values flatten the distribution. The default $\sigma=1.5$ outperforms both $\sigma=1.0$ and $\sigma=3.0$ (Table~\ref{tab:neighborhood_width_ablation}), suggesting that the smoothing distribution should be neither too concentrated nor too flat.

We also compare uniform sampling over $\{6,7,8,9,10\}$, with $\mathbb{E}[K]=8$ and static top-8 inference.
Uniform sampling reaches an Avg(9) score of 30.93, compared with 30.57 for static training and 31.41 for Gaussian 6--10 with $\sigma=1.5$.

\begin{table}[!tbp]
\centering
\begin{tabular}{lcc}
\toprule
\textbf{Training schedule} & $\boldsymbol{\sigma}$ & \textbf{Avg(9)} \\
\midrule
Static top-8 & -- & 30.57 \\
Uniform 6--10 & -- & 30.93 \\
\texttt{6--10} & 1.0 & 31.30 \\
\texttt{6--10} & 1.5 & \textbf{31.41} \\
\texttt{6--10} & 3.0 & 30.66 \\
\bottomrule
\end{tabular}
\caption{
Distribution shape ablation and stochastic-budget controls on OLMoE SFT.
}
\label{tab:neighborhood_width_ablation}
\end{table}

\paragraph{Alternative Budget Controls.}
We compare with controls that increase expert-side computation without smoothing the \texttt{top-8} boundary. Fixed \texttt{top-9} and \texttt{top-10} activate more experts per token during training. The top-$p$ control uses a routing probability threshold of $p=0.4$ and activates 13 experts per token on average, making it the most computationally expensive control. All checkpoints use static top-8
inference unless explicitly stated.

Table~\ref{tab:alternative_budget_controls} shows that static \texttt{top-9},
static \texttt{top-10}, and top-$p$ all remain below \texttt{6--10}, despite
using more expert computation. This suggests that the gain cannot be explained solely by activating more experts, but by smoothing task-loss exposure
across ranks that straddle the deployment cutoff. Full task-level results appear in Tables~\ref{tab:app_sigma_full} (Gaussian shape) and~\ref{tab:app_budget_controls_full} (budget controls).

\begin{table}[!tbp]
\centering
\begin{tabular}{lcc}
\toprule
\textbf{Training rule} & \textbf{Evaluation rule} & \textbf{Avg(9)} \\
\midrule
top-$p$ & static top-8 & 29.72 \\
top-$p$ & top-$p$ & 29.70 \\
\texttt{top-9} & static top-8 & 30.71 \\
\texttt{top-10} & static top-8 & 30.81 \\
\texttt{6--10}, $\sigma=1.5$ & static top-8 & \textbf{31.41} \\
\bottomrule
\end{tabular}
\caption{Alternative budget controls on OLMoE SFT. The final row is the default elastic expert routing strategy from Table~\ref{tab:main_results}.}
\label{tab:alternative_budget_controls}
\end{table}

\section{Conclusion}
\label{sec:conclusion}

In this work, we address the rigid rank-exposure boundary that severely restricts model optimization near the selection threshold under traditional static top-$k$ routing.
We introduce Elastic Expert Routing, which smooths this selection boundary by stochastically sampling from a discrete budget distribution during training while preserving the static top-$k$ inference rule.
We conduct experiments on both supervised fine-tuning and from-scratch pretraining, and the results demonstrate that this simple training-time modification consistently improves MoE performance without increasing inference cost.
Empirical analyses show that elastic training primarily alters the learned routing policy within the foundational middle layers of the network while simultaneously refining expert prioritization rankings near the deployment cutoff. 

\clearpage
\section*{Limitations}
\label{sec:limitations}

The main limitation of this study is the scale of from-scratch pretraining. Our supervised fine-tuning experiments are conducted on established sparse MoE backbones with substantially larger parameter budgets: OLMoE-1B-7B has 7B total parameters with 1B active per token, and Qwen3-30B-A3B has 30.5B total parameters with 3.3B activated. In contrast, our from-scratch pretraining experiments use a 1.427B-parameter MoE with 0.365B active parameters, reflecting the computational cost of training larger MoE models from initialization under available GPU resources. Therefore, the pretraining results should be interpreted as a controlled validation of the proposed training objective at a moderate scale, while validating the approach under large-scale from-scratch pretraining remains an important direction for future work.

\bibliography{references}

\appendix
\setcounter{table}{0}
\renewcommand{\thetable}{\Alph{section}.\arabic{table}}
\setcounter{figure}{0}
\renewcommand{\thefigure}{\Alph{section}.\arabic{figure}}

\section{Experimental Details}
\label{app:reproducibility}

This appendix records the experimental controls behind the main results. The
important design choice is that, within each setting, all routing schedules
share the same model, tokenizer, data, optimizer, evaluation harness, and
decoding protocol. Only the training-time routing schedule changes.

\subsection{Shared Protocol}

Table~\ref{tab:app_shared_protocol} defines the common evaluation surface for
the supervised fine-tuning experiments. Avg(9) is computed over the nine
reported benchmark tasks listed below.

\begin{table}[t]
\centering
\small
\setlength{\tabcolsep}{3pt}
\begin{tabularx}{\linewidth}{@{}p{0.32\linewidth}Y@{}}
\toprule
\textbf{Item} & \textbf{Protocol} \\
\midrule
Schedules &
Static \texttt{top-8}, \texttt{7--9}, \texttt{6--10}, and \texttt{5--11}.
Layer-restricted OLMoE variants apply the dynamic rule only in the specified
layer block. \\
Compute matching &
Elastic schedules are centered at eight experts, preserving the expected active-expert count during training. All main evaluations use static top-8 inference. \\
Main Avg(9) &
MMLU, AGIEval (En), GPQA (Main), BBH, GSM8K, MATH, MBPP, IFEval, and
TruthfulQA. \\
Decoding &
Generation settings follow the task-level lm-evaluation-harness configs. \\

\bottomrule
\end{tabularx}
\caption{Shared evaluation and routing conventions.}
\label{tab:app_shared_protocol}
\end{table}

\subsection{Measured Training Cost}
\label{app:measured_cost}

Table~\ref{tab:app_measured_cost} reports measured training cost for each routing condition. These measurements complement the expected expert-side FLOP
matching used in the main experiments: elastic schedules are centered at eight
experts, while measured wall-clock behavior also depends on dispatcher,
communication, padding, and memory-allocation details.

\begin{table}[t]
\centering
\scriptsize
\setlength{\tabcolsep}{2pt}
\resizebox{\linewidth}{!}{%
\begin{tabular}{llrrr}
\toprule
\textbf{Setting} & \textbf{Method} & \textbf{sec/iter} &
\textbf{Rel.} & \textbf{Max alloc. MB} \\
\midrule
\multicolumn{5}{l}{\textit{OLMoE SFT}} \\
OLMoE & \texttt{top-8} & 3.297 & 1.00x & 91067.81 \\
OLMoE & \texttt{6--10} & 3.394 & 1.03x & 91045.26 \\
OLMoE & \texttt{7--9} & 3.328 & 1.01x & 91044.89 \\
OLMoE & \texttt{5--11} & 3.356 & 1.02x & 91025.12 \\
\midrule
\multicolumn{5}{l}{\textit{Qwen3-30B-A3B SFT}} \\
Qwen3 & \texttt{top-8} & 2.059 & 1.00x & 38729.26 \\
Qwen3 & \texttt{6--10} & 2.167 & 1.05x & 38729.28 \\
Qwen3 & \texttt{7--9} & 2.122 & 1.03x & 38729.28 \\
Qwen3 & \texttt{5--11} & 2.155 & 1.05x & 38729.28 \\
\midrule
\multicolumn{5}{l}{\textit{1.42B MoE Pretraining}} \\
Pretrain & \texttt{top-8} & 2.386 & 1.00x & 45042.56 \\
Pretrain & \texttt{6--10} & 2.242 & 0.94x & 43682.31 \\
Pretrain & \texttt{7--9} & 2.336 & 0.98x & 43216.47 \\
Pretrain & \texttt{5--11} & 2.278 & 0.95x & 43208.01 \\
\bottomrule
\end{tabular}
}
\caption{Measured training cost from the completed training runs. Qwen3-SFT runs use global
batch size 16, micro batch size 1, sequence length 8192, all-to-all token
dispatch, no padding to expert capacity,
and static top-8 inference for all reported evaluations. Pretraining
runs use global batch size 256, micro batch size 4, sequence length 4096, and
25,034 training iterations. Memory is reported as maximum allocated CUDA memory. }
\label{tab:app_measured_cost}
\end{table}

\paragraph{Controlled system profiling.}
Table~\ref{tab:app_system_profile} reports a separate Qwen3-MoE profiling
comparison on 16 GPUs, with the all-to-all dispatcher, expert parallelism,
data, batch size, and sequence length held fixed between static top-8 and
Elastic \texttt{6--10}. Elastic routing increases mean iteration time by
2.6\%, while peak allocated memory and expert-load statistics remain nearly
unchanged. Both runs record the same number of all-to-all events. Cumulative
all-to-all time is lower for elastic routing in this profile, despite its
slightly higher end-to-end iteration time.

\begin{table}[t]
\centering
\footnotesize
\setlength{\tabcolsep}{3pt}
\begin{tabularx}{\linewidth}{@{}Xrr@{}}
\toprule
\textbf{Metric} & \textbf{Static top-8} & \textbf{Elastic 6--10} \\
\midrule
Mean active experts & 8.000 & 7.998 \\
Mean iteration time (s) & 2.204 & 2.261 \\
Median iteration time (s) & 2.214 & 2.234 \\
P95 iteration time (s) & 2.389 & 2.459 \\
Peak allocated memory (MB) & 50,591.09 & 50,591.12 \\
Expert-load CV & 0.4335 & 0.4341 \\
Maximum/mean expert load & 5.6048 & 5.5961 \\
Profiled all-to-all events & 8,640 & 8,640 \\
Cumulative all-to-all time (ms) & 5,657.94 & 3,340.05 \\
Mean GPU utilization while active & 87.7\% & 85.3\% \\
\bottomrule
\end{tabularx}
\caption{Controlled system profiling of Qwen3-MoE on 16 GPUs.
CV denotes the coefficient of variation, and P95 is the 95th percentile.
All-to-all time is the cumulative duration of the recorded events.}
\label{tab:app_system_profile}
\end{table}

\subsection{Supervised Fine-Tuning}

The two supervised fine-tuning settings use different infrastructure because
they start from different sparse MoE backbones. The comparison is therefore
within-backbone: each elastic schedule is compared only with the static top-8
baseline trained under the same recipe.

\paragraph{OLMoE.}
The OLMoE experiments start from OLMoE-1B-7B-0924 and use a standard Hugging Face supervised fine-tuning pipeline.
 We train on Dolci-Instruct-SFT for four epochs. We use AdamW with a peak learning rate
$2{\times}10^{-5}$, linear learning-rate decay, a warmup ratio of 0.03, no weight decay, and
bf16 precision. The global batch size is 256, the sequence length is 4096, and
the runs use 16 GPUs. Router auxiliary and balance coefficients are
both 0.01; the router z-loss coefficient is 0.0. Dynamic budget sampling uses $\sigma=1.5$.

\paragraph{Qwen3.}
The Qwen3 experiments start from Qwen3-30B-A3B-Base
 and use \texttt{ms-swift} \citep{zhao2025swiftascalablelightweightinfrastructure} with
Megatron-LM \citep{shoeybi2020megatronlmtrainingmultibillionparameter}. We train on Infinity-Instruct 7M Core
for three epochs. We use AdamW with $\beta_1{=}0.9$, $\beta_2{=}0.95$, $\epsilon{=}10^{-8}$, weight
decay 0.1, peak learning rate $10^{-5}$, minimum learning rate $10^{-6}$,
warmup fraction 0.05, and bf16 precision. The global batch size is 16 with
micro batch 1, sequence length 8192, PP=2, EP=8, and 16 GPUs. Dynamic budget
sampling uses $\sigma=1.5$; the MoE auxiliary loss coefficient is $10^{-3}$.

\paragraph{Evaluation.}
Both SFT settings are evaluated with lm-evaluation-harness, using the shared Avg(9)
suite in Table~\ref{tab:app_shared_protocol}. 

\subsection{From-Scratch Pretraining}

The pretraining experiment tests whether the same budget-neighborhood effect
appears without supervised fine-tuning. Table~\ref{tab:pretrain_architecture}
summarizes the architecture and data configuration.

\begin{table}[H]
\centering
\small
\setlength{\tabcolsep}{5pt}
\begin{tabular}{lc}
\toprule
\textbf{Item} & \textbf{Value} \\
\midrule
Total parameters & 1.42B \\
Active parameters / token & 0.36B \\
Transformer layers & 20 \\
Hidden size & 768 \\
Attention heads & 24 \\
GQA groups & 6 \\
MoE experts / layer & 128 \\
Activated experts / token & 8 \\
MoE FFN hidden size & 192 \\
Training corpus & FineWeb-Edu \\
Training tokens & 26.25B \\
Validation split & 100M tokens \\
Sequence length & 4096 \\
Training steps & 25,034 \\
\bottomrule
\end{tabular}
\caption{Architecture and data configuration for the from-scratch MoE
pretraining experiment.}
\label{tab:pretrain_architecture}
\end{table}

We set the learning rate and global batch size using the MoE hyperparameter
scaling laws of \citet{tian2025greaterleveragescalinglaws}. Let $M$ denote the
non-embedding forward FLOPs per token and $D$ the number of training tokens, so
the compute budget is $C=M\cdot D$. For our pretraining setting, this gives
$C\approx6.0{\times}10^{19}$. The MoE hyperparameter scaling laws from
\citet{tian2025greaterleveragescalinglaws} are
\[
\begin{aligned}
\eta_{\mathrm{opt}} &= 1.1576 C^{-0.1529},\\
B_{\mathrm{opt}} &= 0.0694 C^{0.3644}.
\end{aligned}
\]
We therefore use learning rate
$1.09516{\times}10^{-3}$ and global batch size 256 with sequence length 4096,
corresponding to 1.05M tokens per update.

All variants use Megatron Adam with a constant learning-rate schedule, 251
warmup steps, weight decay 0.01, micro batch size 4, and bf16 precision. The
routing conditions are static top-8 and the elastic schedules \texttt{7--9},
\texttt{6--10}, and \texttt{5--11}; dynamic budget sampling uses
$\sigma=1.5$. System settings are TP=1, PP=1, EP=4 with an all-to-all
dispatcher on 16 GPUs.

\Needspace{9\baselineskip}
\subsection{Additional Baseline Comparisons}
\label{app:training_method_comparison}

We provide the baseline configurations and task-level results underlying
the training-method comparison in Section~\ref{sec:training_method_comparison}.
The comparisons cover Qwen3-30B-A3B SFT and 1.42B MoE pretraining, with
separate results for each setting.

\paragraph{Baseline configurations.}
We use Noisy top-$k$ with noise scale $\sigma=0.25$ and expert FFN dropout with probability $p=0.40$. Noisy top-$k$ perturbs the routing scores while retaining a fixed expert count; FFN dropout is applied to expert activations. DefaultMoE uses moving-average defaults with $\beta=0.999$ for uncomputed expert outputs to provide dense router feedback \citep{panda2025defaultmoe}. Elastic routing uses the \texttt{6--10} neighborhood with $\sigma=1.5$.

All methods follow the corresponding Qwen3 SFT or 1.42B pretraining recipe in Appendix~\ref{app:reproducibility}, using the same backbone, training data, and optimization budget. Evaluation uses static top-8 routing and the task suites of Tables~\ref{tab:main_results} and~\ref{tab:pretrain_validation}. Tables~\ref{tab:baseline_tasks_sft} and~\ref{tab:baseline_tasks_pt} report the complete task-level results.
The SFT and pretraining averages cover nine and eleven tasks, respectively.
LAMBADA perplexity is reported separately and is excluded from the
pretraining average, following Table~\ref{tab:training_method_comparison}.

\begin{table*}[t]
\centering
\small
\setlength{\tabcolsep}{6pt}
\begin{tabular}{lrrrrr}
\toprule
\textbf{Task} & \textbf{Static top-8} & \textbf{Noisy top-$k$} & \textbf{FFN dropout} & \textbf{DefaultMoE} & \textbf{Elastic 6--10} \\
\midrule
MMLU & 78.24 & 78.61 & 77.35 & 78.13 & 78.49 \\
AGIEval (En) & 51.74 & 52.23 & 51.90 & 52.19 & 52.28 \\
GPQA (Main) & 37.72 & 40.18 & 36.83 & 38.51 & 39.06 \\
BBH & 49.29 & 45.74 & 64.80 & 57.87 & 63.29 \\
GSM8K & 81.43 & 84.61 & 81.91 & 82.70 & 82.03 \\
Minerva Math & 41.24 & 42.30 & 42.18 & 42.35 & 42.36 \\
MBPP & 73.80 & 73.00 & 70.20 & 72.09 & 74.00 \\
IFEval & 41.22 & 45.29 & 42.03 & 42.26 & 42.51 \\
TruthfulQA & 53.14 & 51.31 & 52.25 & 52.57 & 51.93 \\
\midrule
Avg(9) & 56.42 & 57.03 & 57.72 & 57.63 & 58.44 \\
\bottomrule
\end{tabular}
\caption{Task-level comparison of MoE training methods on Qwen3-30B-A3B SFT. Avg(9) is the unweighted average over the nine tasks. Minerva Math corresponds to MATH in Table~\ref{tab:main_results}.}
\label{tab:baseline_tasks_sft}
\end{table*}

\begin{table*}[t]
\centering
\small
\setlength{\tabcolsep}{6pt}
\begin{tabular}{lrrrrr}
\toprule
\textbf{Task} & \textbf{Static top-8} & \textbf{Noisy top-$k$} & \textbf{FFN dropout} & \textbf{DefaultMoE} & \textbf{Elastic 6--10} \\
\midrule
LAMBADA PPL $\downarrow$ & 54.2 & 45.6 & 54.1 & 48.2 & 42.7 \\
LAMBADA & 31.3 & 30.5 & 31.1 & 31.8 & 33.1 \\
HellaSwag & 41.9 & 43.0 & 38.4 & 42.5 & 43.7 \\
PIQA & 68.4 & 67.8 & 66.0 & 66.8 & 68.6 \\
WinoGrande & 49.3 & 51.1 & 51.2 & 51.3 & 52.6 \\
CommonsenseQA & 19.4 & 20.3 & 20.1 & 19.7 & 20.1 \\
SciQ & 81.4 & 81.2 & 70.3 & 80.4 & 82.0 \\
OpenBookQA & 31.0 & 31.8 & 33.2 & 32.6 & 33.8 \\
ARC-Easy & 53.7 & 55.5 & 51.6 & 55.3 & 55.1 \\
ARC-Challenge & 28.5 & 29.4 & 28.2 & 29.2 & 29.9 \\
MMLU & 23.1 & 24.8 & 23.9 & 25.5 & 26.1 \\
LogiQA & 26.3 & 28.1 & 27.2 & 27.5 & 27.2 \\
\midrule
Average (excl. PPL) & 41.3 & 42.1 & 40.1 & 42.0 & 42.9 \\
\bottomrule
\end{tabular}
\caption{Task-level comparison of MoE training methods in 1.42B MoE pretraining. All scores except LAMBADA perplexity are percentages; the average excludes perplexity.}
\label{tab:baseline_tasks_pt}
\end{table*}

\Needspace{12\baselineskip}
\subsection{Full Task-Level Ablation Results}
\label{app:full_ablation_results}

Tables~\ref{tab:app_sigma_full} and~\ref{tab:app_budget_controls_full} provide
the task-level OLMoE results for the ablations in
Section~\ref{sec:ablations_controls}. Table~\ref{tab:app_sigma_full} varies
the Gaussian scale while keeping the \texttt{6--10} support fixed.
Table~\ref{tab:app_budget_controls_full} compares top-$p$, fixed
\texttt{top-9}, and fixed \texttt{top-10} training with the default elastic
schedule. All comparisons use static top-8 evaluation, except for the
additional top-$p$ evaluation column. Avg(9) uses the same nine-task
unweighted average as the main SFT results.

\begin{table}[t]
\centering
\small
\setlength{\tabcolsep}{4.5pt}
\begin{tabular}{lccc}
\toprule
\textbf{Task} & $\boldsymbol{\sigma=1.0}$ & $\boldsymbol{\sigma=1.5}$ & $\boldsymbol{\sigma=3.0}$ \\
\midrule
MMLU & 50.65 & 51.25 & 50.65 \\
AGIEval & 23.62 & 23.96 & 23.31 \\
GPQA & 27.90 & 26.34 & 27.68 \\
BBH & 35.68 & 35.28 & 35.23 \\
GSM8K & 19.79 & 21.46 & 19.26 \\
MATH & 6.30 & 6.44 & 5.70 \\
MBPP & 27.60 & 25.40 & 25.00 \\
IFEval & 49.35 & 51.02 & 47.69 \\
TruthfulQA & 40.85 & 41.58 & 41.41 \\
\midrule
Avg(9) & 31.30 & \textbf{31.41} & 30.66 \\
\bottomrule
\end{tabular}
\caption{Full task-level results for the distribution-shape ablation in
Table~\ref{tab:neighborhood_width_ablation}. All rows use the same
\texttt{6--10} budget support and static top-8 evaluation; only the Gaussian
scale $\sigma$ changes.}
\label{tab:app_sigma_full}
\end{table}

\begin{table}[t]
\centering
\small
\setlength{\tabcolsep}{3.2pt}
\resizebox{\columnwidth}{!}{%
\begin{tabular}{lccccc}
\toprule
\textbf{Task} & \textbf{top-$p$/8} & \textbf{top-$p$/top-$p$} & \textbf{\texttt{top-9}} & \textbf{\texttt{top-10}} & \textbf{\texttt{6--10}} \\
\midrule
MMLU & 50.06 & 50.06 & 50.66 & 50.25 & 51.25 \\
AGIEval & 23.44 & 23.44 & 23.36 & 23.29 & 23.96 \\
GPQA & 27.90 & 27.90 & 26.34 & 26.12 & 26.34 \\
BBH & 34.00 & 34.00 & 33.71 & 35.16 & 35.28 \\
GSM8K & 17.29 & 17.29 & 19.33 & 20.32 & 21.46 \\
MATH & 4.72 & 4.72 & 6.22 & 7.04 & 6.44 \\
MBPP & 24.80 & 24.80 & 25.00 & 25.40 & 25.40 \\
IFEval & 45.10 & 44.92 & 50.28 & 49.54 & 51.02 \\
TruthfulQA & 40.21 & 40.21 & 41.46 & 40.19 & 41.58 \\
\midrule
Avg(9) & 29.72 & 29.70 & 30.71 & 30.81 & \textbf{31.41} \\
\bottomrule
\end{tabular}
}
\caption{Full task-level results for the alternative budget controls in
Table~\ref{tab:alternative_budget_controls}. All models use static top-8
evaluation unless otherwise stated; top-$p$/8 denotes top-$p$ training with
static top-8 evaluation.}
\label{tab:app_budget_controls_full}
\end{table}

\end{document}